\documentclass{article}
\usepackage{ijcai22}

\usepackage{times}

\usepackage{soul}
\usepackage{url}
\usepackage[hidelinks]{hyperref}
\usepackage[utf8]{inputenc}
\usepackage[small]{caption}
\usepackage{graphicx}
\usepackage{amsmath}
\usepackage{booktabs}
\usepackage{algorithm}
\usepackage{algorithmic}
\usepackage{float}
\usepackage[utf8]{inputenc} 
\usepackage[T1]{fontenc}    
\usepackage{hyperref}       
\usepackage{url}            
\usepackage{booktabs}       
\usepackage{amsfonts}       
\usepackage{nicefrac}       
\usepackage{microtype}      
\usepackage{xcolor,amsmath}         
\usepackage{algorithm}
\usepackage{algorithmic}
\usepackage{graphicx}
\title{Stochastic Weight Averaging Revisited}

\author{
Hao Guo\and
Jiyong Jin\and
Bin Liu\footnote{Contact Author}\\
\affiliations
Research Center for Applied Mathematics and Machine Intelligence\\
Zhejiang Lab, Hangzhou, China\\
\emails
\{guoh, jinjy, liubin\}@zhejianglab.com
}

\begin{document}

\maketitle

\begin{abstract}
Averaging neural network weights sampled by a backbone stochastic gradient descent (SGD) is a simple yet effective approach to assist the backbone SGD in finding better optima, in terms of generalization. From a statistical perspective, weight averaging (WA) contributes to variance reduction. Recently, a well-established stochastic weight averaging (SWA) method is proposed, which is featured by the application of a cyclical or high constant (CHC) learning rate schedule (LRS) in generating weight samples for WA. Then a new insight on WA appears, which states that WA helps to discover wider optima and then leads to better generalization. We conduct extensive experimental studies for SWA, involving a dozen modern DNN model structures and a dozen benchmark open-source image, graph, and text datasets. We disentangle contributions of the WA operation and the CHC LRS for SWA, showing that the WA operation in SWA still contributes to variance reduction but does not always lead to wide optima. The experimental results indicate that there are global scale geometric structures in the DNN loss landscape. We then present an algorithm termed periodic SWA (PSWA) which makes use of a series of WA operations to discover the global geometric structures. PSWA outperforms its backbone SGD remarkably, providing experimental evidences for the existence of global geometric structures. Codes for reproducing the experimental results are available at \url{https://github.com/ZJLAB-AMMI/PSWA}.
\end{abstract}
\begin{algorithm}[t]
\caption{Stochastic Weight Averaging (SWA)}
\label{alg:swa}
\textbf{Input}: weights $w_{\tiny{\mbox{SGD}}}$, LRS, cycle length $c$, number of iterations $n$\\
\textbf{Output}: $w_{\tiny{\mbox{SWA}}}$
\begin{algorithmic}[1] 
\STATE $w\leftarrow w_{\tiny{\mbox{SGD}}}$; $w_{\tiny{\mbox{SWA}}}\leftarrow w$.
\FOR{$i\leftarrow 1,2,\ldots,n$}
\STATE Compute current learning rate $\alpha$ according to the LRS.
\STATE $w\leftarrow w-\alpha\bigtriangledown \mathcal{L}_i(w)$ (stochastic gradient update).
\IF {mod($i$,$c$)=0}
\STATE $n_{\tiny{\mbox{models}}}\leftarrow i/c$ (number of models averaged).
\STATE $w_{\tiny{\mbox{SWA}}}\leftarrow \left(w_{\tiny{\mbox{SWA}}}\cdot n_{\tiny{\mbox{models}}}+w\right)/\left(n_{\tiny{\mbox{models}}}+1\right)$.
\ENDIF
\ENDFOR
\STATE \textbf{return} $w_{\tiny{\mbox{SWA}}}$
\end{algorithmic}
\end{algorithm}
\section{Introduction}
Stochastic gradient descent (SGD) equipped with a decaying learning rate schedule (LRS) is the \emph{de facto} approach to train modern deep neural networks (DNNs). Averaging neural network (NN) weights sampled by a backbone SGD is shown to be a simple yet effective approach to assist the backbone SGD in finding better optima, in terms of generalization. The idea of weight averaging (WA), also referred to as iterate averaging or tail-averaging \cite{jain2018parallelizing}, goes back to \cite{ruppert1988efficient,polyak1992acceleration}. A WA procedure averages the final few iterates of SGD. From a statistical perspective, it has been proved that, the WA operation contributes to decreasing the variance in the final iterate of its backbone SGD, resulting in a stabilizing effect in terms of regularization properties and prediction guarantees \cite{neu2018iterate}. We term this view as variance reduction in what follows.

Recently, a stochastic WA (SWA) method has been proposed and received a lot of attentions. It is extremely easy to implement yet could improve SGD to achieve better generalization without the significant computational overhead \cite{izmailov2018averaging,cha2021swad,hwang2021adversarial}. SWA starts after a converged SGD (namely the backbone SGD), which runs preceding it and outputs a local optimum $w_{\tiny{\mbox{SGD}}}$ of the loss function $f(w)$, where $w$ denotes the NN weights. SWA rewarms its backbone SGD, starting at $w_{\tiny{\mbox{SGD}}}$. The rewarmed SGD employs a cyclical or high constant (CHC) LRS. The application of the CHC LRS is the major feature that discriminates SWA from other WA methods. Novel local optima are sampled along the trajectory of this rewarmed SGD process. Then a WA operation is used, which outputs the mean of such optima, denoted by $w_{\tiny{\mbox{SWA}}}$, as the final output of SWA. A pseudo-code to implement SWA is shown in Algorithm 1, which outputs  a running average of the sampled weights per $c$ iterations.

A common insight to explain SWA's success is that the local optima discovered by its rewarmed backbone SGD are located at the boundary of a high-quality basin region in the DNN weight parameter space. Doing WA over such local optima then results in a wider optimum, which is closer to the center of the basin region \cite{izmailov2018averaging}, and a wider optimum leads to better generalization \cite{izmailov2018averaging,keskar2016large}.

We now have two seemingly independent views on the role of WA, one is statistical, namely the variance reduction perspective, and the other is geometric, namely the wider optimum perspective. Then, what is the relationship between these two views? and, how do they reconcile?

After a detailed inspection of SWA \cite{izmailov2018averaging}, we find that its behavior results from a combined effect of several possible intertwined factors, namely the convergence rate of the SGD that runs preceding SWA, the CRC LRS, the WA operation, and finally the application of the momentum technique and weight decaying. The common geometric view can not explain the specific role of each factor. For example, it can not answer the following questions:
\begin{enumerate}
  \item if we do not use the momentum technique in its backbone SGD, how does SWA behave?
  \item if the SGD that runs preceding SWA does not converge or converges to a bad optimum, can SWA still work?
  \item what is the actual function of the WA operation to SWA, variance reduction or discovering wider optimum or both?
\end{enumerate}

The above concerns motivate us to revisit SWA. As SWA is a fundamental, generic, architecture-agnostic technique for training DNNs, any new findings, insight from this re-inspection could bring a broad potential impact on deep learning.  The major contributions of this paper can be summarized as follows,
\begin{enumerate}
  \item we disentangle contributions of the WA operation, the CHC LRS, the application of momentum and weight decaying, and the rate at which the preceding SGD converges, to the behavior of SWA.
  \item we find that the actual function of the WA operation in SWA is variance reduction, in the same spirit as tail-averaging \cite{jain2018parallelizing}.
  \item we find cases in which SWA fails to discover better optima than its backbone SGD.
  \item we find experimental evidence for the existence of global geometric structures in the DNN loss landscape; we show that such global structures can be exploited by the WA operation.
  \item we propose a novel algorithm design termed periodic SWA (PSWA) inspired by the above experimental finding, and demonstrate that it is preferable to SGD when the training budget is so limited that it can not support an SGD to converge.
\end{enumerate}
\section{Related Work}
\textbf{Iterates Averaging}
The basic idea of iterate averaging, also referred to as tail-averaging in \cite{jain2018parallelizing}, goes back to \cite{ruppert1988efficient,polyak1992acceleration}. The tail-averaging method averages the final few iterates of SGD. In this way, it decreases the variance in the final iterate of SGD and brings a stabilizing effect in terms of regularization properties and prediction guarantees \cite{neu2018iterate}. A generalization error bounds for tail-averaging in the context of least square regression with the stochastic approximation is derived in \cite{jain2018parallelizing}. We show in this paper that the WA operation is a type of tail-averaging, which gives the stabilizing effect and the variance reduction function to SWA.

\textbf{Cyclical Learning Rates}
The benefits of employing cyclical learning rates (CLRs) in SGD have been demonstrated in \cite{smith2017cyclical,loshchilov2016sgdr}. Following that, such CLR strategy has been widely used in developing advanced DNN optimizers, such as fast geometric ensembling (FGE) \cite{garipov2018loss}, snapshot ensembles \cite{huang2017snapshot}, super-convergence training \cite{smith2019super}, or exploring the loss landscape of DNNs \cite{smith2017exploring}. In this paper, we characterize that the CLR strategy also plays a major role in SWA's success.

\textbf{Convergence Theory}
In \cite{allen2019convergence}, Zhu et al. present a convergence theory for training DNNs, based on two assumptions: the input data points are distinct and the DNN architecture is over-parameterized. This theory tells that, at least for fully-connected neural networks (NNs), convolutional NNs (CNN), and residual NNs (ResNet), SGD with a random weight initialization can attain 100\% accuracy in classification tasks with the number of iterations scaling polynomial in the number of training samples and the number of NN layers. Cheridito et al. demonstrate that, for ReLU networks that have a much larger depth than their width, SGD fails to converge if the number of restarted SGD trajectories does not increase to infinity fast enough \cite{cheridito2021non}. Here we investigate specific roles of the CHC LRS and the WA operation in promoting SGD's convergence. While our result is empirical, it may stimulate more theoretical research on DNN convergence.

\textbf{Loss Landscape Study \& Sharpness-aware Minimization}
Another commonly used way to investigate the convergence problem is through loss landscape analysis. The Hessian spectrum analysis has shown to be an effective approach to inspect smoothness, curvature, and sharpness of NN loss landscapes \cite{ghorbani2019investigation,sagun2017empirical}. Yao et al. developed an open-source scalable framework for fast computation of Hessian information in DNNs \cite{yao2020pyhessian}. It has been common wisdom that, at least for some cases, NNs generalize better when they converge to a wider local optimum, and vice versa \cite{keskar2016large}. However, the correlation between the local sharpness of the loss landscape and the global property like generalization performance may be only correlative, other than causative \cite{yang2021taxonomizing}.

The empirical finding of the relationship between local sharpness and global generalization motivated the design of practical approaches to improving the generalization property of SGD. For example, the sharpness-aware minimization (SAM) method seeks NN weights that lie in a wider loss basin by modifying the optimization objective function to be sharpness-aware \cite{kleiner2021sharpness}. SWA can be seen as a type of wideness-aware solver for DNN optimization. It is reported that SWA finds wider minima than SAM \cite{cha2021swad}. Our work characterizes the root cause that leads to SWA's success and provides more empirical evidence for deeply understanding the loss landscape of DNNs.
\section{Main Results}\label{sec:main}
Our goal is to inspect the real cause that leads to SWA's behavior. Toward this goal, we experiment with different DNN architectures on different datasets. We present the main results indexed with questions of our interest. All details about the experimental settings are described in Section \ref{sec:experiment_setting}.
\subsection{Does SWA always find wider optima than SGD?}\label{sec:does}
The results reported in the SWA paper \cite{izmailov2018averaging} show that SGD generally converges to a boundary of a wide basin region and SWA helps to find an optimum exactly located in that wide basin region. All experiments conducted there use image datasets, such as CIFAR-$\{10,100\}$ \cite{krizhevsky2009learning}, and ImageNet ILSVRC-2012 \cite{deng2009imagenet,russakovsky2015imagenet}. We wonder whether SWA always finds wider optima than SGD. We conduct experiments on graph and text datasets. Results show that the answer is no. Specifically, on a graph dataset MUTAG, we use SWA to train a graph isomorphism network (GIN) for graph classification. The baseline optimizer selected is Adam, which is an advanced SGD method that performs better for graph data based tasks. We find that if we run Adam with 300 epochs, then we get a test accuracy (TA) value 89\%, while if we replace Adam with SWA for the last 30 epochs, we can only get a smaller TA value 84\%.

We consider the graph node classification task using graph neural network (GNN) models, such as graph convolutional network (GCN) \cite{yao2019graph}, GraphSAGE \cite{hamilton2017inductive}, and graph attention network (GAT) \cite{velivckovic2017graph}, using public open-source datasets Cora, Citeseer, and Pubmed. The parameter setting for the experiment is shown in Table \ref{tab:parameter_GNN_exp} in Section \ref{sec:exp_graph}. The TA comparison result is presented in Table \ref{tab:res_GNN_exp}.
\begin{table*}
\caption{Test accuracy (\%) comparison between Adam and SWA for the graph node classification task on datasets Cora, Citeseer and Pubmed. Best results are \textbf{bolded}.}
\label{tab:res_GNN_exp}
\centering
\begin{tabular}{c|c|c|c} 
\hline
 & Cora & Citeseer  & Pubmed \\
\begin{tabular}{c} 
 \\ \hline GCN   \\  GraphSAGE  \\  GAT
\end{tabular} 
& \begin{tabular}{cc} 
Adam & SWA \\ \hline
81.2$_{\pm0.47}$ & \textbf{81.3}$_{\pm0.21}$\\
\textbf{80.2}$_{\pm0.22}$ & 79.3$_{\pm0.49}$\\
81.6$_{\pm0.43}$ & \textbf{81.7}$_{\pm0.52}$\\
\end{tabular} 
& \begin{tabular}{cc} 
Adam & SWA \\ \hline
70.5$_{\pm0.45}$ & \textbf{70.8}$_{\pm0.41}$\\
\textbf{69.8}$_{\pm0.54}$ & 69.7$_{\pm0.16}$\\
\textbf{70.6}$_{\pm0.26}$ & 70.4$_{\pm0.25}$\\
\end{tabular}  
& \begin{tabular}{cc} 
Adam & SWA \\ \hline
79.5$_{\pm0.29}$ & \textbf{79.6}$_{\pm0.26}$\\
\textbf{77.8}$_{\pm0.17}$ & 77.7$_{\pm0.29}$\\
76.1$_{\pm0.43}$ & \textbf{76.2}$_{\pm0.28}$\\
\end{tabular} 
\\ \hline
\end{tabular} 
\end{table*}

We also consider a graph classification task using models MinCutPool \cite{bianchi2020spectral} and SAGPool \cite{lee2019self}, on public open-source datasets NCI1 (\url{https://paperswithcode.com/dataset/nci1}), D\&D (\url{https://paperswithcode.com/sota/graph-classification-on-dd}), and PROTEINS (\url{https://paperswithcode.com/sota/graph-classification-on-proteins}). The parameter settings are shown in Tables \ref{tab:parameter_GNN_exp_NCI1}-\ref{tab:parameter_GNN_exp_PROTEINS} in Section \ref{sec:exp_graph}. The TA comparison result is shown in Table \ref{tab:res_GNN_exp_all}.
\begin{table*}
\caption{Test accuracy (\%) comparison between Adam and SWA for the graph classification task on datasets NCI1, D\&D, and PROTEINS. Best results are \textbf{bolded}.}
\label{tab:res_GNN_exp_all}
\centering
\begin{tabular}{c|c|c|c} 
\hline
 & NCI1 & D\&D  & PROTEINS \\
\begin{tabular}{c} 
 \\ \hline MinCutPool   \\  SAGPool
\end{tabular} 
& \begin{tabular}{cc} 
Adam & SWA \\ \hline
74.78$_{\pm0.42}$ & \textbf{75.25$_{\pm0.14}$}\\
71.63$_{\pm0.88}$ & \textbf{72.57$_{\pm0.49}$}\\
\end{tabular} 
& \begin{tabular}{cc} 
Adam & SWA \\ \hline
79.67$_{\pm0.69}$ & \textbf{80.86$_{\pm0.85}$}\\
\textbf{71.89$_{\pm0.33}$} & 70.87$_{\pm0.45}$\\
\end{tabular}  
& \begin{tabular}{cc} 
Adam & SWA \\ \hline
76.44$_{\pm1.50}$ & \textbf{77.34$_{\pm1.83}$}\\
77.73$_{\pm1.16}$ & \textbf{78.37$_{\pm1.10}$}\\
\end{tabular} 
\\ \hline
\end{tabular} 
\end{table*}

The above experimental results for graph datasets show that using SWA does not always lead to better generalization than some advanced SGD optimizers like Adam.

On a text dataset termed Microsoft Research Paraphrase Corpus (MRPC), we use an SGD with momentum to fine-tune the pre-trained model RoBERTa for testing whether two sentences are semantically equivalent. See details about the experimental setting in Section \ref{sec:experiment_setting_1}. On average, SGD could give a TA value 87.98\%, while SWA only achieves 87.50\%.

We find that, even for image datasets, SWA does not always converge to a boundary of a wide basin region, especially when we remove the momentum module from its backbone SGD. For such cases, we find that SWA may converge to a deep loss valley, where the averaged gradients over mini-batch training samples are all close to zero. Then the products of such gradients and the learning rate are close to zero. In such cases, SWA fails to find a wider optimum with better generalization. See details of the experimental results in Section \ref{sec:exp_clean}.
\subsection{What is the real function of the WA operation to SWA?}\label{sec:role}
To answer the question in the title of this section, we conduct ablation studies across different DNN models and datasets. As mentioned above, the SWA procedure consists of a CHC LRS based rewarmed SGD process, from which a set of NN weights are sampled, and a WA operation that yields the average of these sampled weights. We refer to the sampled weights as SWA samples in what follows. The momentum and weight decaying operations are removed here to provide a clean investigation.

\begin{figure*}
\centering
\includegraphics[width=0.34\linewidth]{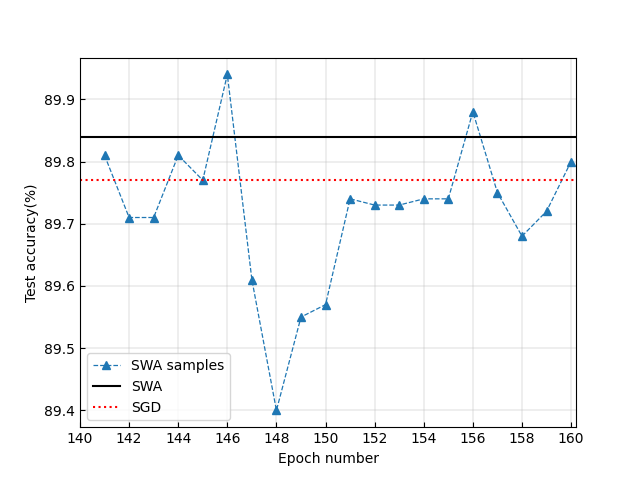}\includegraphics[width=0.34\linewidth]{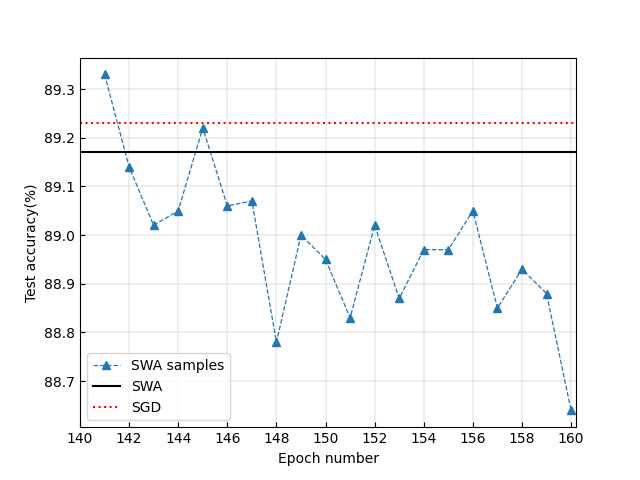}\includegraphics[width=0.34\linewidth]{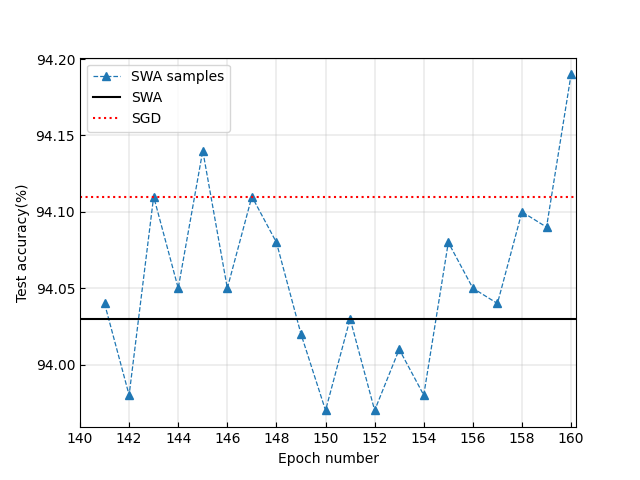}\\
\includegraphics[width=0.34\linewidth]{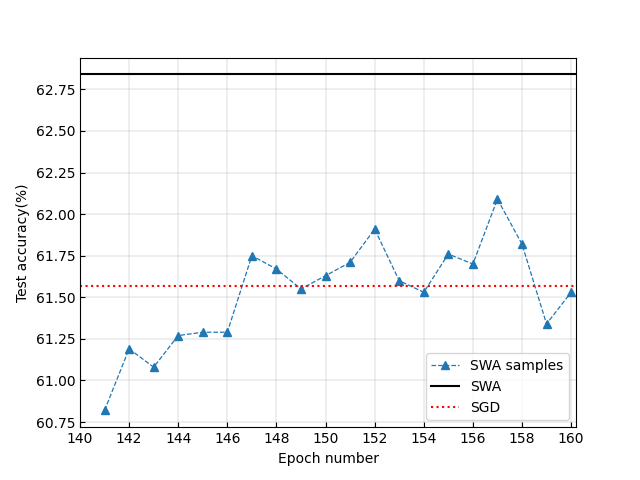}\includegraphics[width=0.34\linewidth]{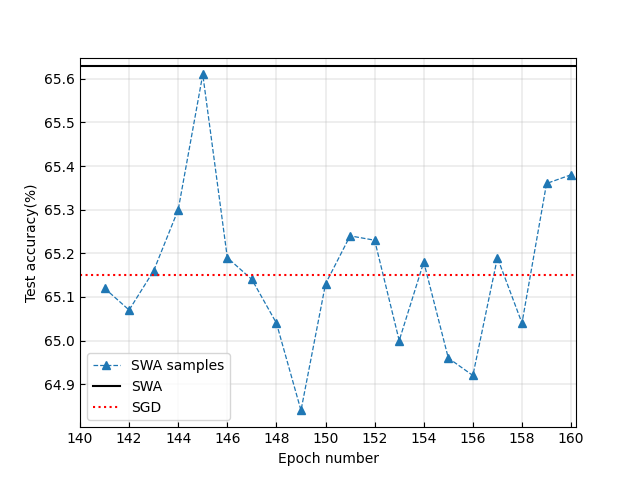}\includegraphics[width=0.34\linewidth]{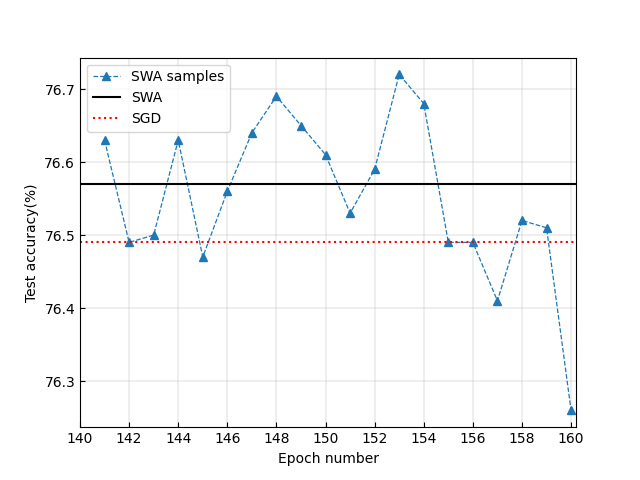}
\caption{Ablation study of the CHC LRS and the WA operation for DNNs that converge well. The legend ``SGD" denotes the TA value associated with the NN weight given by the backbone SGD at the time point when SWA is started. The legend ``SWA samples" denotes TA values associated with NN weights sampled during the SWA procedure. The legend ``SWA" denotes the TA value associated with the mean of NN weights sampled during the SWA procedure. The sub-figures in the left/middle/right column correspond to VGG16/PreResNet-164/WideResNet-28-10. The sub-figures in the top/bottom row correspond to dataset CIFAR-10/100.}\label{fig:ablation}
\end{figure*}

First, we consider image classification with DNN structures VGG16 \cite{simonyan2014very}, Preactivation ResNet-164 (PreResNet-164) \cite{he2016identity}, WideResNet-28-10 \cite{zagoruyko2016wide}, using datasets CIFAR-$\{$10,100$\}$. For each model, SWA runs after a preceding SGD process being converged. The results are presented in Figure \ref{fig:ablation}. The effect of using CHC LRS can be revealed by comparing TA values of SWA samples to that of SGD. The effect of WA can be checked by comparing the TA value of SWA and those of separate SWA samples. As is shown, neither using the CHC LRS nor performing WA brings a significantly clear benefit for increasing the TA value. This result coincides with that revealed in the above subsection.

\begin{figure*}
\centering
\includegraphics[width=0.34\linewidth]{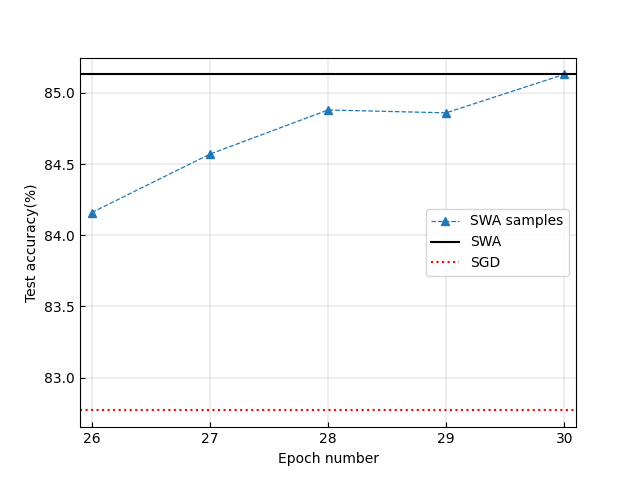}\includegraphics[width=0.34\linewidth]{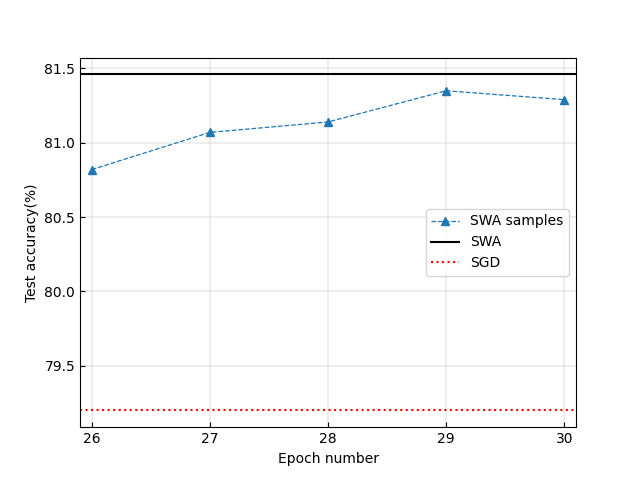}\includegraphics[width=0.34\linewidth]{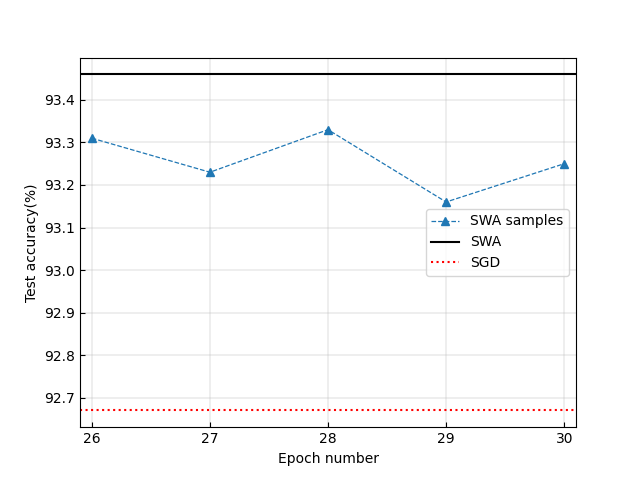}\\
\includegraphics[width=0.34\linewidth]{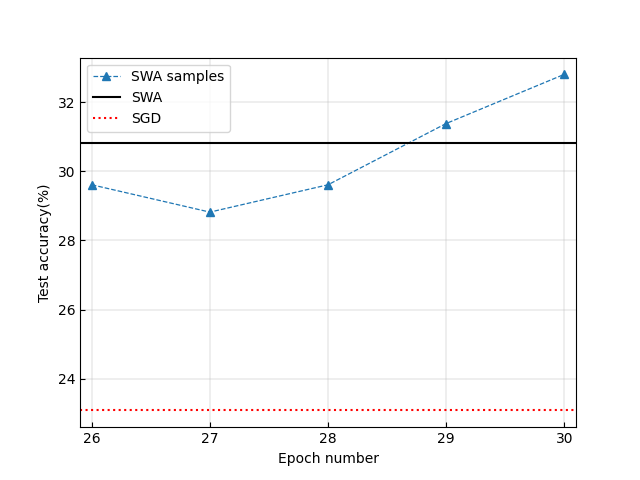}\includegraphics[width=0.34\linewidth]{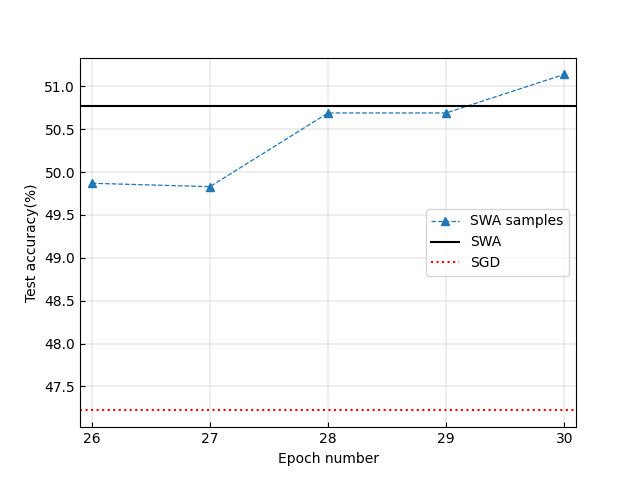}\includegraphics[width=0.34\linewidth]{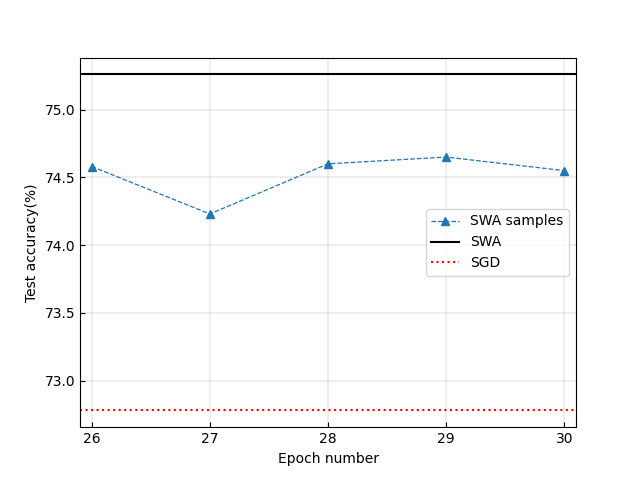}
\caption{Ablation study on the CHC LRS and the WA operation for DNNs that does not converge well. The legends are defined in the same way as in Figure \ref{fig:ablation}. The sub-figures in the left/middle/right column correspond to VGG16/PreResNet-164/WideResNet-28-10. The sub-figures in the top/bottom row correspond to dataset CIFAR-10/CIFAR-100.}\label{fig:ablation2}
\end{figure*}

We then consider cases in which the backbone SGD that runs preceding SWA converges to a bad optimum, corresponding to Case II in Section \ref{sec:experiment_setting_2}. In this case, we do not give enough budget for DNN training. The number of training epochs is only 30. The results based on CIFAR-$\{$10,100$\}$ are presented in Figure \ref{fig:ablation2}. In this case, it is shown that, the application of the CHC LRS makes the resulting SWA samples produce striking greater TA values than the backbone SGD that runs before SWA. We also find that WA contributes an additional increase in the TA value.

Finally, we conduct an ablation study based on the Imagenet dataset. See the result in Section \ref{sec:exp_imagenet}. It is shown that SWA samples provide much bigger TA values than SGD; and, except for VGG16, the WA operation provides an additional increase in the TA value.

For all cases aforementioned, the WA operation always outputs a TA value that is bigger than the smallest TA value given by separate SWA samples. That says the WA operation in SWA functions similarly as tail-averaging \cite{jain2018parallelizing}, in decreasing the variance of TA values associated with SWA samples.
\section{Periodic SWA}\label{sec:pswa}
\subsection{On global geometric structure of the DNN loss landscape}\label{sec:insight}
As presented above, we find cases in which SWA is initialized by a backbone SGD that does not converge well and SWA performs strikingly better than its backbone SGD, while if the backbone SGD converges well, then the performance gap of SWA and SGD is reduced or even becomes indistinguishable. As we know, when the backbone SGD converges well, then the NN weights employed by SWA shall center around a local optimum discovered by this SGD. Thus, SWA can only make use of a very local geometric structure around this local optimum. When the backbone SGD does not converge well, then NN samples fed to SWA shall span a much wider area. This motivates us to raise a hypothesis as follows:

Is there any global geometric structure in the DNN loss landscape that can be encountered by an SGD at the early stage of its life cycle? If such a global structure exists, how to exploit it for facilitating the discovery of higher quality local optima?

We propose a novel algorithm design, termed periodic SWA (PSWA) that starts at an early stage of its backbone SGD. PSWA exploits the aforementioned possible global structures via performing WA sequentially. We show experimental results in the following section, which demonstrate that PSWA outperforms its backbone SWA remarkably, thus provides evidence for the existence of such global geometric structures.

PSWA consists of a series of SWA procedures that run sequentially. The first SWA procedure is initialized by an NN weight given by the backbone SGD that runs preceding SWA. For each of the other SWA procedures, its starting weight seed is the output of its former SWA procedure. Different from the original SWA method, which is invoked when its preceding SGD is converged, PSWA is started when its preceding SGD is at a very early stage of its working period. In addition, PSWA uses a LRS that is totally the same as its backbone SGD, as shown in Figure \ref{fig:lrs}. If the sequentially performed SWA procedures can continually bring performance gains compared with the backbone SGD, then it would indicate that PSWA has made use of some global structures of the loss landscape to search local optima.

Notably, the PSWA algorithm is a byproduct of our experimental findings in Section \ref{sec:main}. The aim of our experiments here is to test the whether our hypothesis raised in subSection \ref{sec:insight} holds.
\subsection{On performance of PSWA}
PSWA consists of a series of SWA procedures that run sequentially. The first SWA procedure is initialized by an NN weight given by the backbone SGD that runs preceding SWA. For each of the other SWA procedures, its starting weight seed is the output of its former SWA procedure. Different from the original SWA method, which is invoked when its preceding SGD is converged, PSWA is started when its preceding SGD is at a very early stage of its working period. In addition, PSWA uses a LRS that is totally the same as its backbone SGD, as shown in Figure \ref{fig:lrs}. If the sequentially performed SWA procedures can continually bring performance gains compared with the backbone SGD, then it would indicate that PSWA has made use of some global structures of the loss landscape to search local optima.

Notably, the PSWA algorithm is a byproduct of our experimental findings in Section \ref{sec:main}. The aim of our experiments here is to test the whether our hypothesis raised in subSection \ref{sec:insight} holds.

We compare PSWA with the backbone SGD on datasets CIFAR-10 and CIFAR-100, based on DNN structures VGG16, PreResNet-164, and WideResNet-28-10. The momentum factor for SGD is 0.9, and the weight decaying parameter is set at 0.0005. PSWA starts after the 40th epoch with a period of 20 epochs. Within one period of PSWA, a full SWA procedure is conducted. In a SWA procedure, we sample one NN weight per epoch, then average the weights that have been sampled within this SWA procedure as the current output of PSWA.
 \begin{figure*}
\centering
\includegraphics[width=0.34\linewidth]{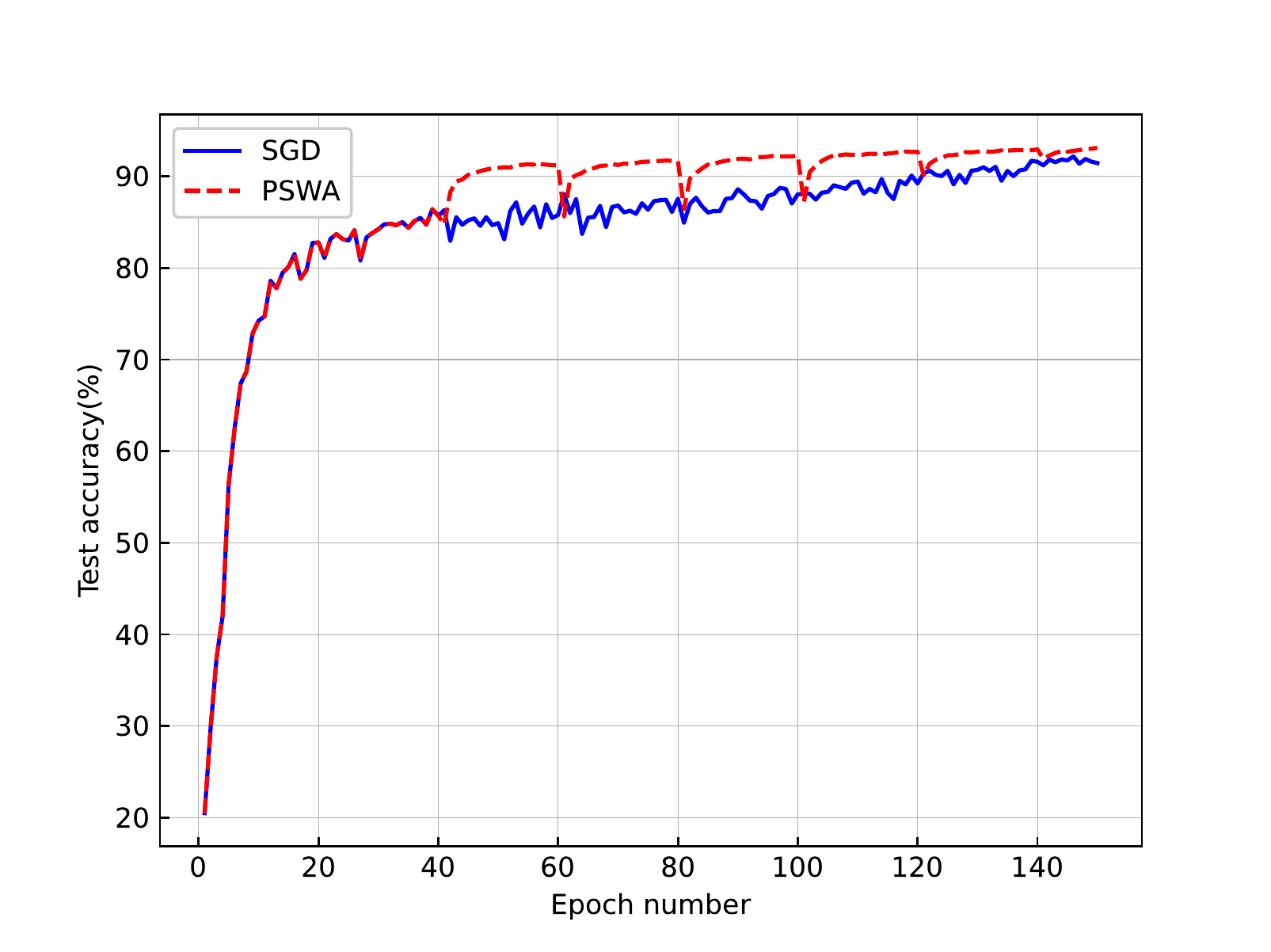}\includegraphics[width=0.34\linewidth]{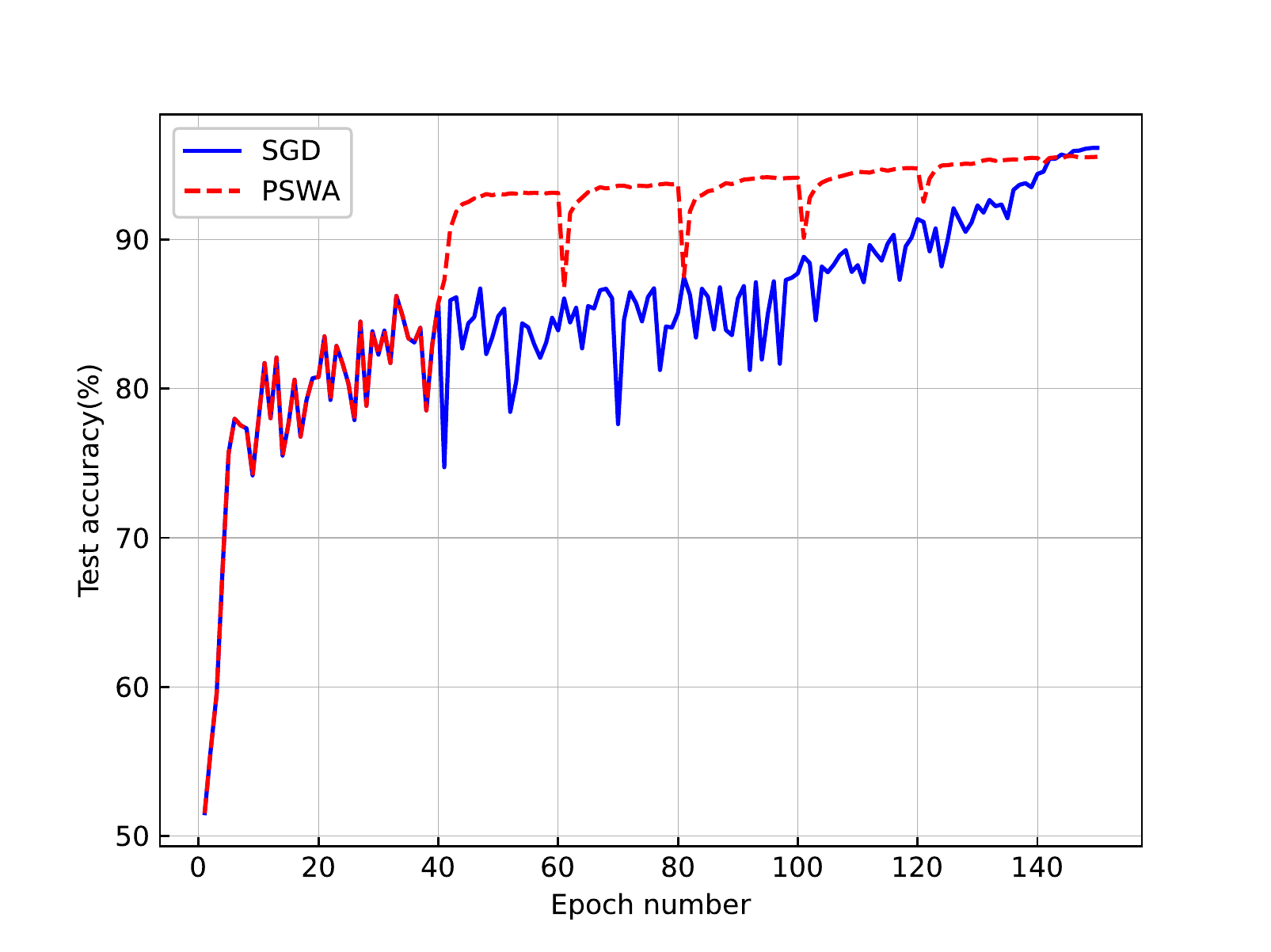}\includegraphics[width=0.34\linewidth]{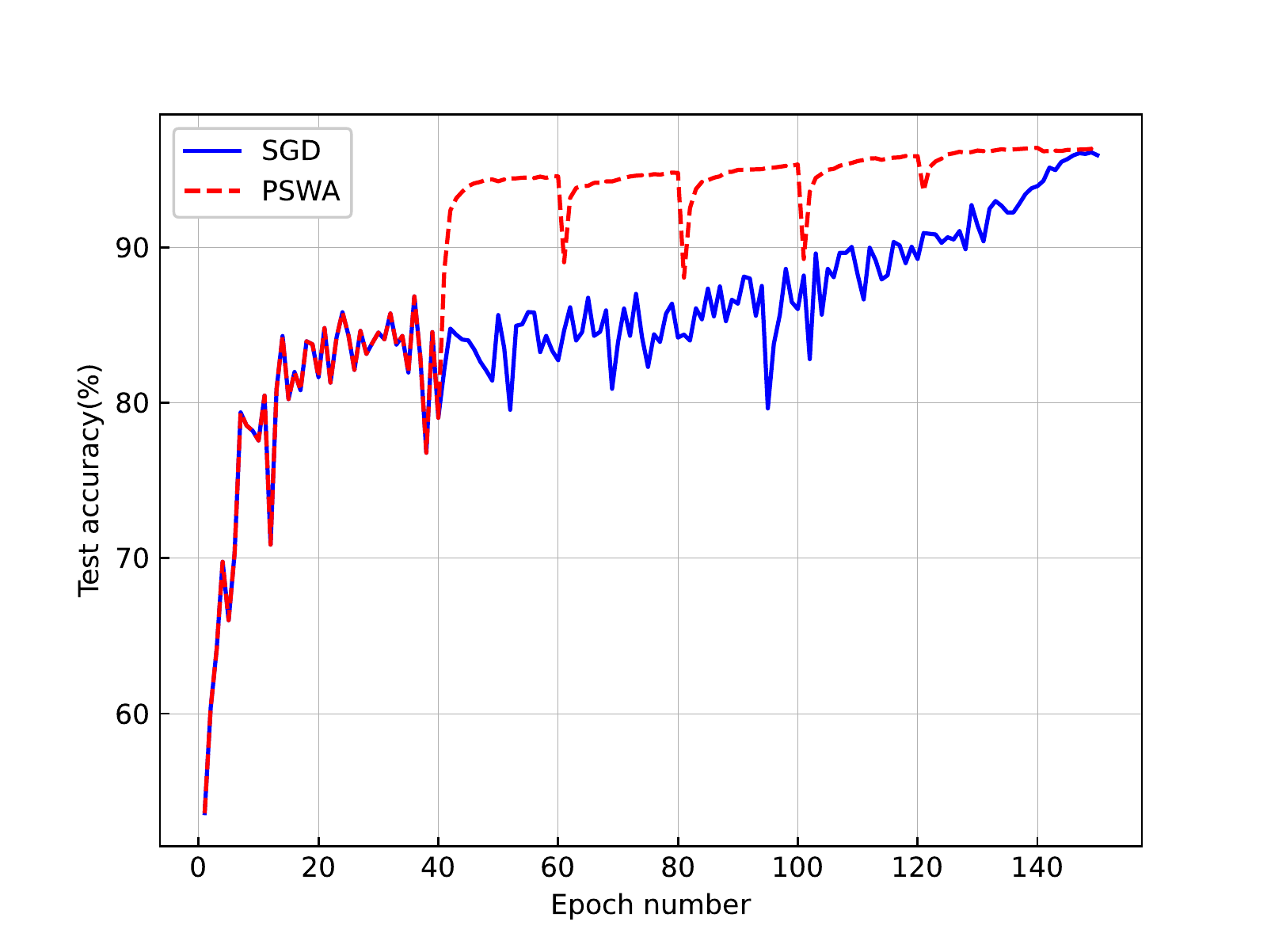}\\
\includegraphics[width=0.34\linewidth]{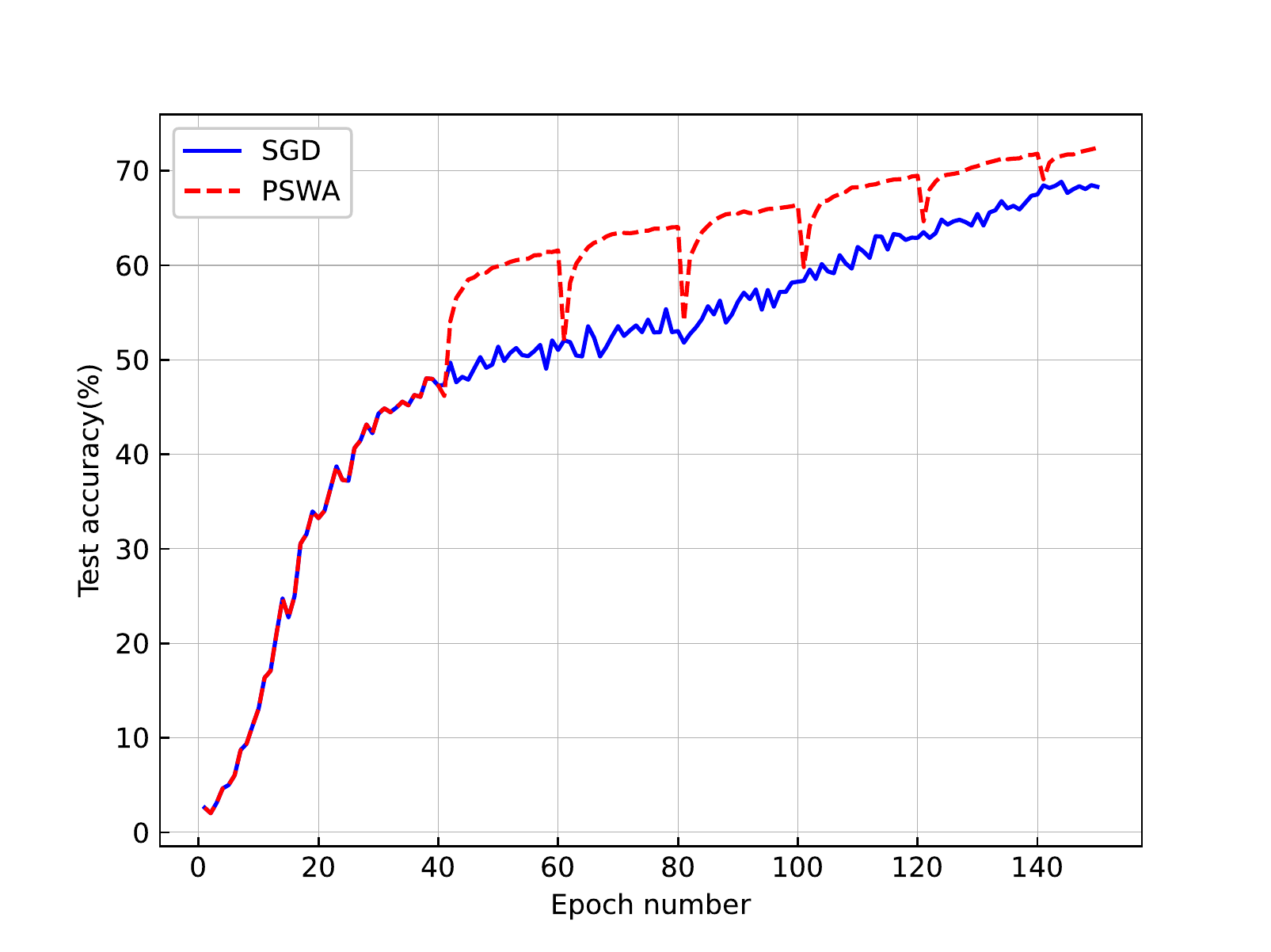}\includegraphics[width=0.34\linewidth]{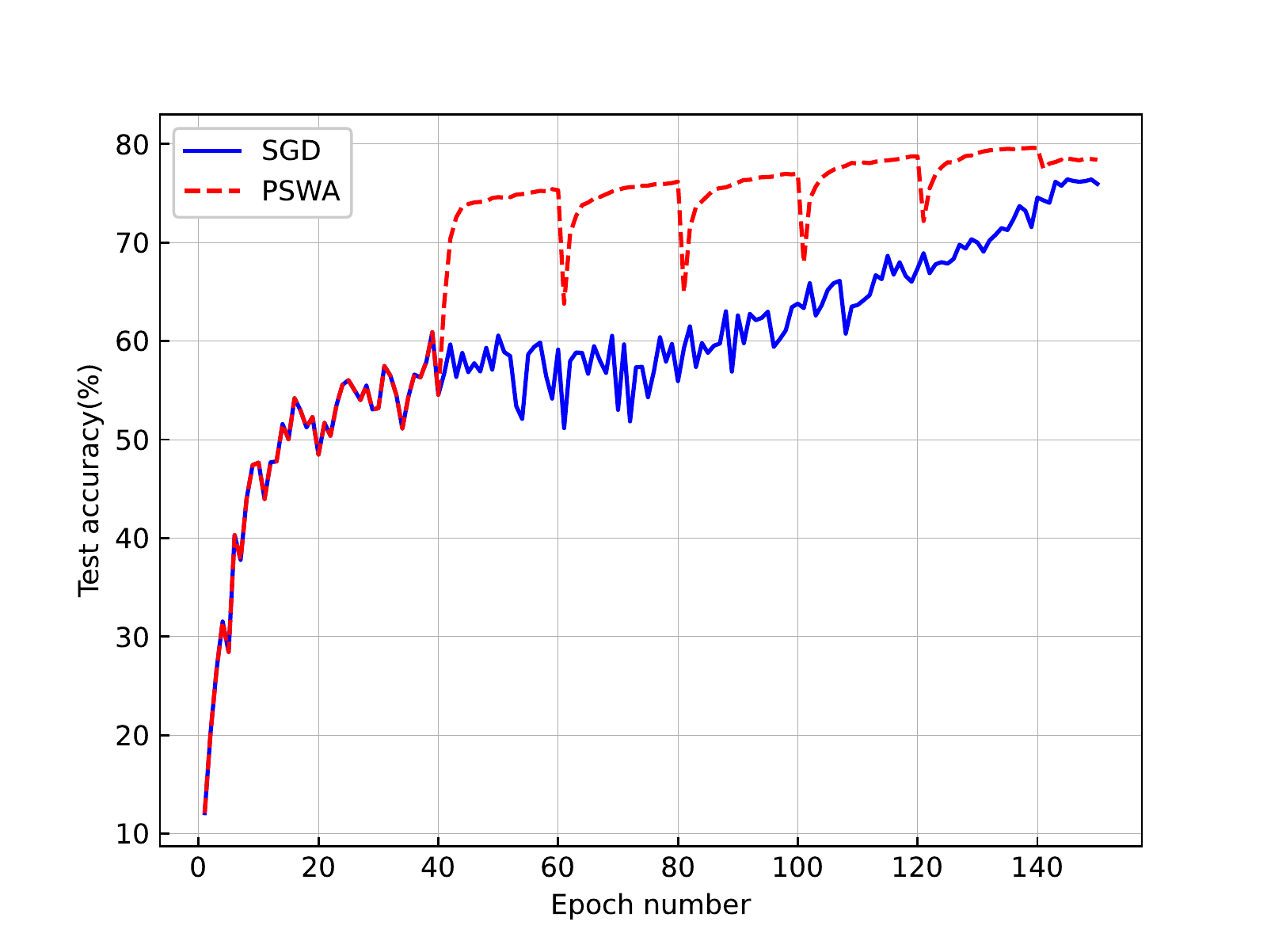}\includegraphics[width=0.34\linewidth]{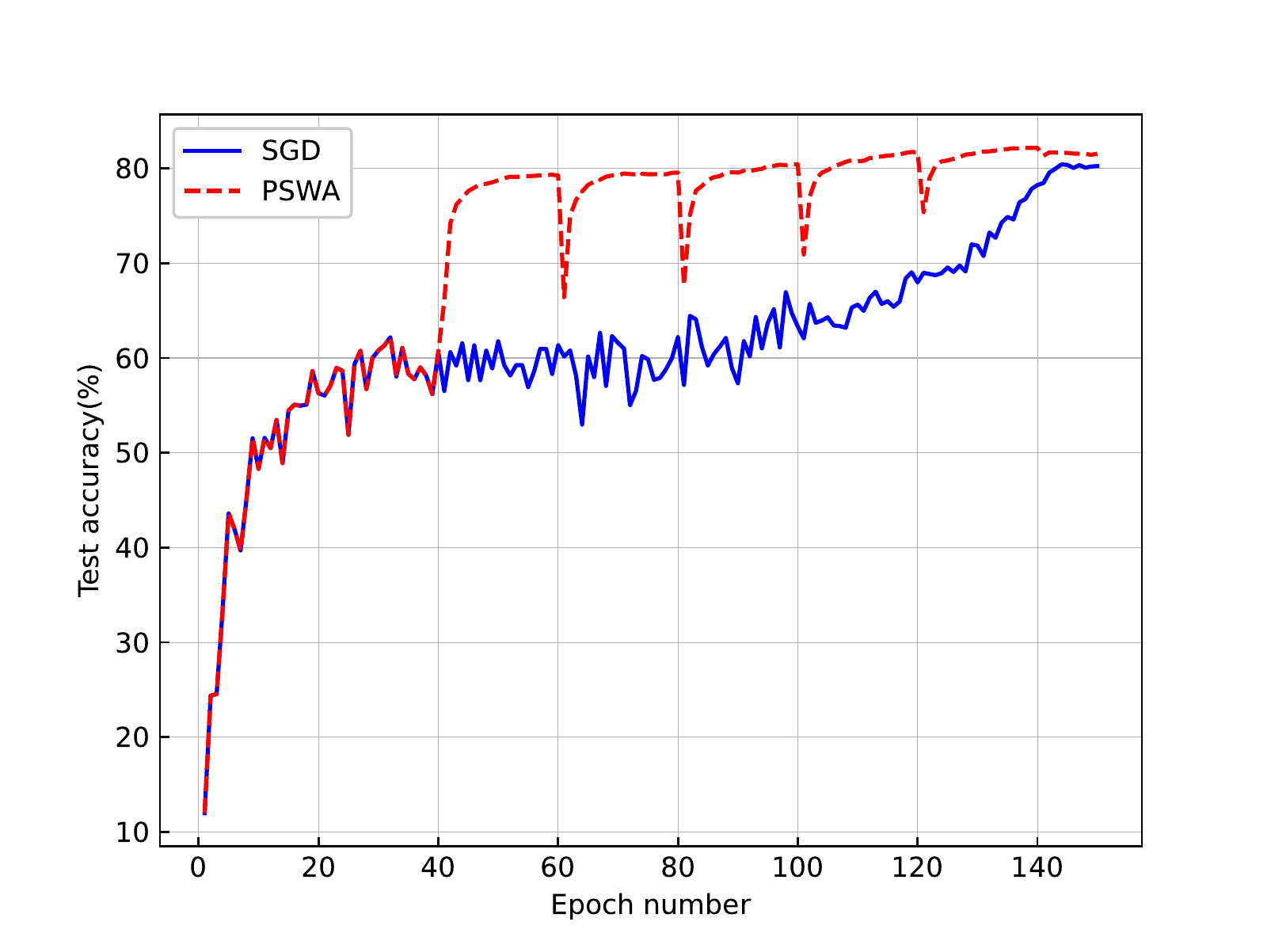}
\caption{Test accuracy comparison between PSWA and its backbone SGD. The sub-figures in the left/middle/right column correspond to VGG16/PreResNet-164/WideResNet-28-10. The sub-figures in the top/bottom row correspond to dataset CIFAR-10/CIFAR-100.}\label{fig:pswa}
\end{figure*}

The experimental results are shown in Fig.\ref{fig:pswa}. We see that PSWA indeed provides a remarkable performance gain compared with its backbone SGD at the early stage of the training process. This provides experimental evidence for the existence of global geometric structures in the DNN loss landscape that can be encountered by an SGD process at the early stage of its working period, and demonstrates that such structures can be exploited by the WA operations for improving the backbone SGD. From an algorithmic perspective, we could not claim that PSWA is better than SGD, since qualities of their final outputs at the end of the training process are indistinguishable, while if the training budget can not support the whole process of training, then PSWA is clearly preferable to SGD, since it provides much better weight samples than SGD at the early stage of the training process.

We also compare two special examples of PSWA, termed double SWA (DSWA) and triple SWA (TSWA), to SWA. DSWA and TSWA consist of two and three sequentially performed SWA procedures, respectively. See the pseudo-codes to implement DSWA and TSWA in Algorithm \ref{alg:dswa} and Algorithm \ref{alg:tswa}. To make a fair comparison, we let SWA, DSWA, and TSWA run the same number of iterations to guarantee that their computational budgets are almost the same. We do not use the momentum and weight decaying, to get rid of their influences on the comparison.
\begin{algorithm}[!htbp]
\caption{Double Stochastic Weight Averaging (DSWA)}
\label{alg:dswa}
\textbf{Input}: weights $\hat{w}$, LRS, cycle length $c$, number of iterations $n$ (assumed to be multiples of 2)\\
\textbf{Output}: $w_{\mbox{dswa}}$
\begin{algorithmic}[1] 
\STATE Run the SWA procedure (namely Algorithm 1) with input $\hat{w}$, $c$, $n/2$. Denote the output to be $w_{\mbox{swa}}$.
\STATE $\hat{w}\leftarrow w_{\mbox{swa}}$.
\STATE Run the SWA procedure again with input $\hat{w}$, $c$, $n/2$. Denote the output to be $w_{\mbox{dswa}}$.
\STATE \textbf{return} $w_{\mbox{dswa}}$
\end{algorithmic}
\end{algorithm}
\begin{algorithm}[!htbp]
\caption{Triple Stochastic Weight Averaging (TSWA)}
\label{alg:tswa}
\textbf{Input}: weights $\hat{w}$, LRS, cycle length $c$, number of iterations $n$ (assumed to be multiples of 3)\\
\textbf{Output}: $w_{\mbox{tswa}}$
\begin{algorithmic}[1] 
\STATE Run the SWA procedure (namely Algorithm 1) with input $\hat{w}$, $c$, $n/3$. Denote the output to be $w_{\mbox{swa}}$.
\STATE $\hat{w}\leftarrow w_{\mbox{swa}}$.
\STATE Run the SWA procedure again with input $\hat{w}$, $c$, $n/3$. Denote the output to be $w_{\mbox{dswa}}$.
\STATE $\hat{w}\leftarrow w_{\mbox{dswa}}$.
\STATE Run the SWA procedure again with input $\hat{w}$, $c$, $n/3$. Denote the output to be $w_{\mbox{tswa}}$.
\STATE \textbf{return} $w_{\mbox{tswa}}$
\end{algorithmic}
\end{algorithm}
\begin{table}[!htbp]
\caption{Test accuracy (\%) comparison among SGD, SWA, and DSWA on CIFAR-10 based on a toy CNN model. The preceding SGD procedure does not converge. The best results is \textbf{bolded}.}
\centering
\begin{tabular}{llll}
\hline
SGD & SWA & DSWA \\
\hline
57.10${_{\pm0.48}}$ & 67.27${_{\pm0.29}}$ & \textbf{69.49${_{\pm0.33}}$} \\
\hline
\end{tabular}
\label{tab:TA_Converge}
\end{table}
\begin{table}[!htbp]
\caption{Test accuracy (\%) comparison among SGD, SWA, DSWA, and TSWA on CIFAR-100. The SGD procedure that runs before SWA does not converge. Best results are \textbf{bolded}.}
\centering
\begin{tabular}{cccc}
\hline
& VGG16 & PreResNet-164 & WideResNet-28-10 \\
\hline
SGD & 55.28$_{\pm0.62}$ & 70.55$_{\pm0.84}$  & 76.30$_{\pm0.81}$ \\
SWA & 65.89$_{\pm0.24}$ & 76.45$_{\pm0.63}$  & 80.95$_{\pm0.27}$ \\
DSWA &68.44$_{\pm0.25}$ & 77.26$_{\pm0.49}$ & \textbf{81.18}$_{\pm0.14}$   \\
TSWA & \textbf{68.68}$_{\pm0.16}$ & \textbf{77.33}$_{\pm0.45}$  & 81.11$_{\pm0.12}$    \\
\hline
\end{tabular}
\label{tab:SWAs_comparison}
\end{table}

\begin{figure}[!htbp]
\centering
\includegraphics[width=0.5\linewidth]{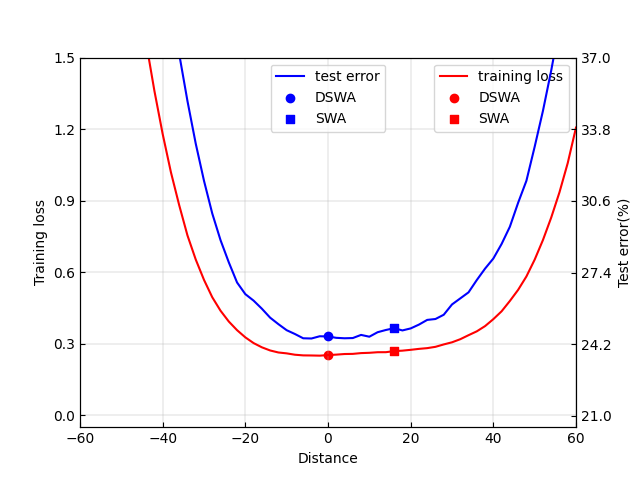}\includegraphics[width=0.5\linewidth]{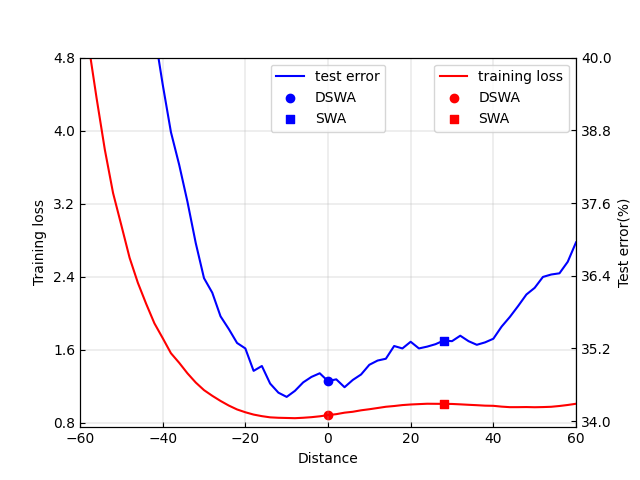}\\
\includegraphics[width=0.5\linewidth]{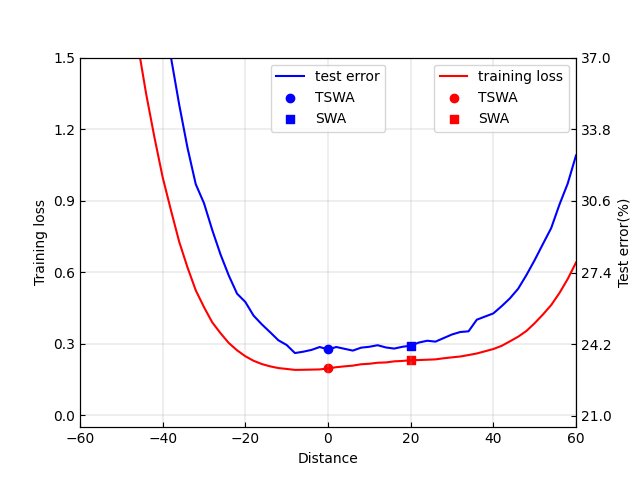}\includegraphics[width=0.5\linewidth]{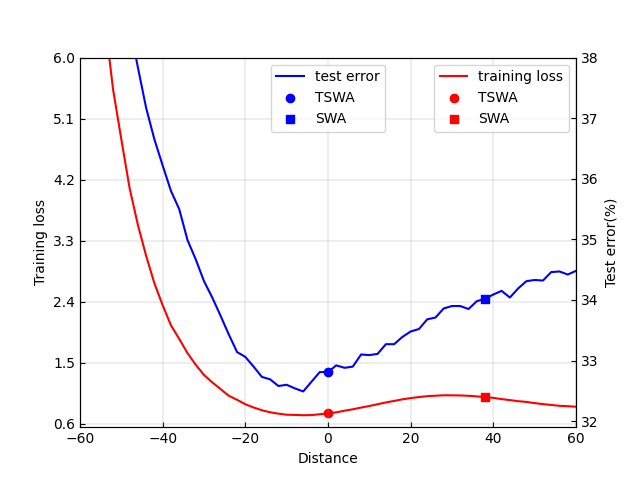}
\caption{Cross-entropy train loss and test error as a function of a point on the line connecting SWA
and DSWA (or TSWA) solutions on CIFAR-100. DSWA and TSWA are initialized by a non-converged preceding SGD procedure. Left: PreResNet-164. Right: VGG16.}\label{fig:compare_dswa}
\end{figure}

\begin{figure}[!htbp]
\centering
\includegraphics[width=0.5\linewidth]{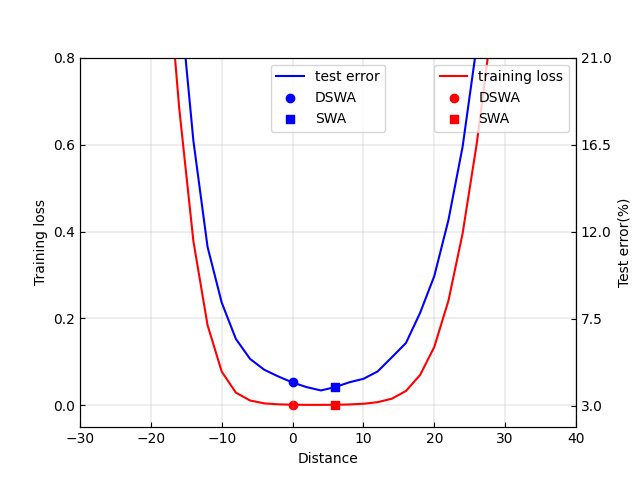}\includegraphics[width=0.5\linewidth]{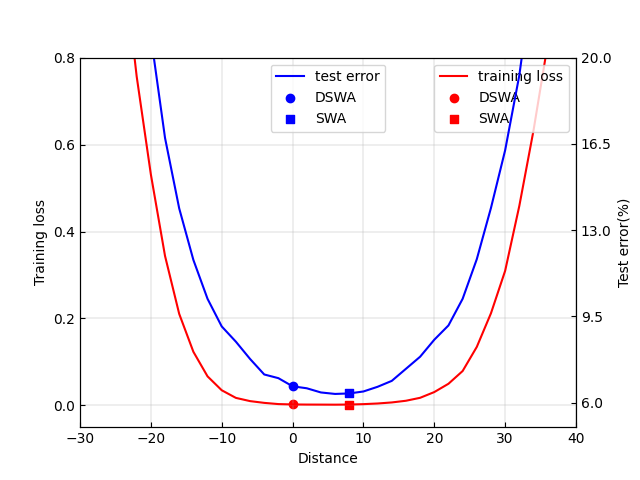}\\
\includegraphics[width=0.5\linewidth]{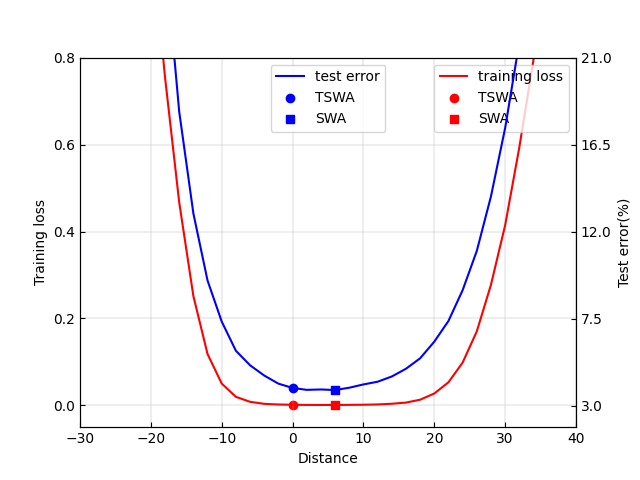}\includegraphics[width=0.5\linewidth]{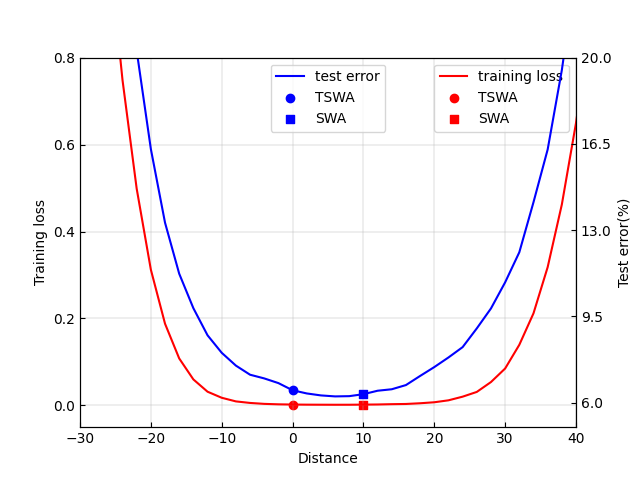}
\caption{Cross-entropy train loss and test error as a function of a point on the line connecting SWA
and DSWA (or TSWA) solutions on CIFAR-10. DSWA and TSWA are initialized by a converged preceding SGD procedure. Left: PreResNet-164. Right: VGG16.}\label{fig:compare_dswa2}
\end{figure}

We find that, if the backbone SGD that runs preceding SWA is non-converged or converges to a bad local optimum, corresponding to Case II in Section \ref{sec:experiment_setting_2}, DSWA and TSWA indeed find flatter optima that lead to better generalization than SWA, see results in Tables \ref{tab:TA_Converge}, \ref{tab:SWAs_comparison} and Figure \ref{fig:compare_dswa}.
If the backbone SGD converges well, corresponding to Case I in Section \ref{sec:experiment_setting_2}, then DSWA and TSWA fail to find flatter optima than SWA, as shown in Figure \ref{fig:compare_dswa2}. Note that Figures \ref{fig:compare_dswa} and \ref{fig:compare_dswa2} are obtained in the same way as that used to obtain Figure 5 in \cite{izmailov2018averaging}.
\section{Conclusions}
In this paper, we investigated how the weight averaging operation and the cyclical or high constant learning rate scheduling
each contribute to SWA. Through experiments on a broad range of NN architectures, we identified a link between SGD and the global loss
landscape and developed a novel insight from a statistical as well as geometric perspective in regard to SWA. Specifically, we find that SWA works because it provides a mechanism to combine advantages of the WA operation and the CHC LRS. The CHC LRS contributes to discovering global scale geometric structures, and WA contributes to exploiting such structures. By leveraging SGD's early training phase behavior, we proposed a novel algorithm, periodic SWA, which is shown to be capable of finding high quality local optima much more quickly than SGD.

Although we cover a broad range of network architectures and different types of datasets in our experiments, our findings still lack theoretical support and may not always hold for all DNN tasks. Hopefully, our work could stimulate more theoretical and algorithmic research on demonstrating, discovering, and exploiting non-local geometric structures of DNN's loss landscape in the future.
\section*{Acknowledgment}
This work was supported by the Research Initiation Project of Zhejiang Lab (No.2021KB0PI01).
\bibliographystyle{IEEEtran} 
\bibliography{ijcai22}
\section{Appendix}
\subsection{Experimental Setting}\label{sec:experiment_setting}
In this section, we describe our experimental setting corresponding to results presented in Sections \ref{sec:main} and \ref{sec:pswa}.
\subsubsection{Experiment setting for results reported in Section \ref{sec:does}}\label{sec:experiment_setting_1}
For the graph classification task, we ran our experiments on a public open-source dataset MUTAG, which is commonly used for the graph classification task. See details about this dataset at \url{https://paperswithcode.com/dataset/mutag}. We use Adam \cite{kingma2014adam} to train a GIN model for 300 epochs. We set the learning rate $\alpha$ at 0.01, and use the default parameter setting for the exponential decay rates $\beta_1$ and $\beta_2$, namely let $\beta_1=0.9$ and $\beta_2=0.999$. For SWA, it starts at the 270th epoch, using a constant learning rate 0.02.

For experiments on the text dataset MRPC (see details about this dataset at \url{https://paperswithcode.com/dataset/mrpc}), the learning rate of SGD is fixed at $10^{-4}$ during the first 20 epochs, then is linearly decreased to $10^{-6}$ in the following 20 epochs, then is fixed at $10^{-6}$ for the last 10 epochs. The momentum and the weight decaying factors are set at 0.9 and 0.01, respectively. SWA is started at the 45th epoch. For each epoch of SWA, the learning rate is linearly decreased from $5\times 10^{-5}$ to $5\times 10^{-6}$.
\subsubsection{Experiment setting for results reported in Section \ref{sec:role}}\label{sec:experiment_setting_2}
\begin{figure}[!htbp]
\centering
\includegraphics[width=0.5\linewidth]{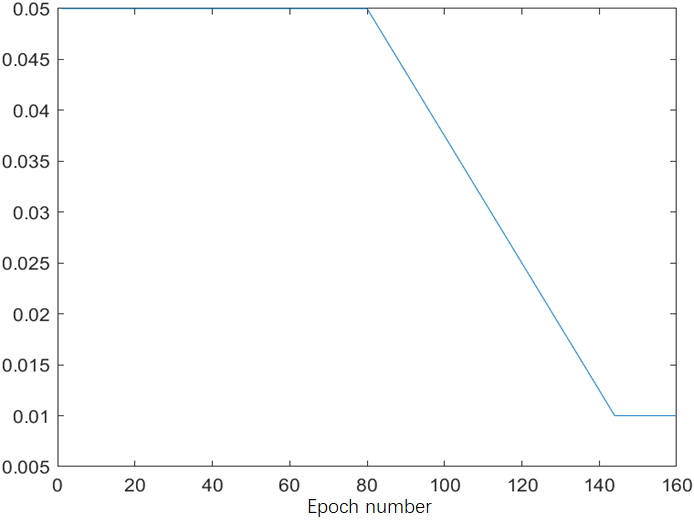}
\caption{An example of the learning rate schedules used in our experiments.}\label{fig:lrs}
\end{figure}

For the image classification experiments presented in Section \ref{sec:role}, we consider two major cases, termed Case I and Case II here, for each DNN architecture under consideration. In Case I, we run SWA after a converged SGD. In Case II, we run it after a non-converged SGD. We adopt the same type of LRS as used in \cite{izmailov2018averaging}. An example of LRS we use is shown in Figure \ref{fig:lrs}. This LRS covers $L=160$ epochs in total. The first half segment of this LRS takes a constant higher value $C_h$, followed by a segment of LRS that consists of linearly decreased learning rate values. The ending segment of this LRS takes a constant lower value $C_l$. For the LRS shown in Figure \ref{fig:lrs}, $C_h=0.05$, and $C_l=0.01$. For Case I, we set the value of $L$ to be big enough, and that of $C_l$ small enough to guarantee that the SGD process that runs before SWA is converged. For Case II, we set a small value like 30 to $L$, to ensure that the SGD that runs preceding SWA does not converge. For the SWA procedure, the cycle length $c$ takes a value that makes a cycle equal to an epoch. We adopt the same CHC LRS as used in \cite{izmailov2018averaging} for the SWA procedure. The mini-batch size is set at 128 for all experiments.
\subsection{Experiment on Imagenet}\label{sec:exp_imagenet}
\begin{figure}[!htbp]
\centering
\includegraphics[width=0.5\linewidth]{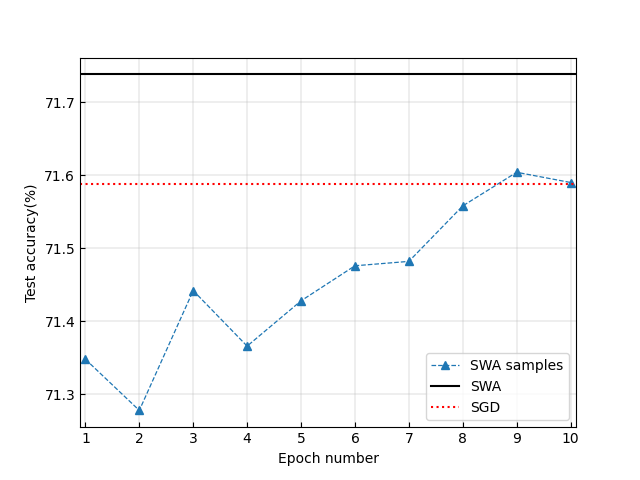}\includegraphics[width=0.5\linewidth]{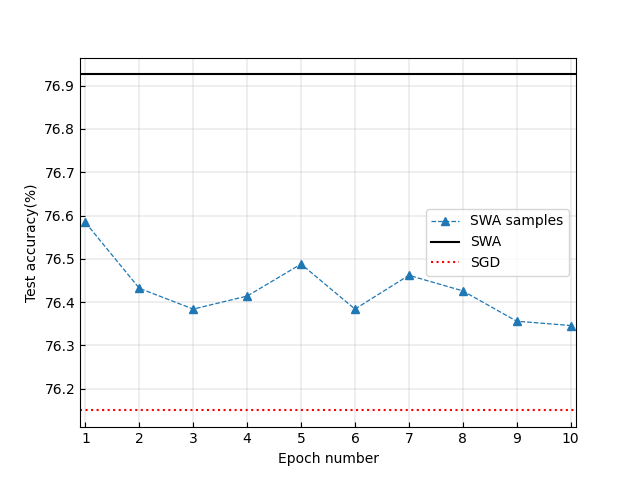}\\
\includegraphics[width=0.5\linewidth]{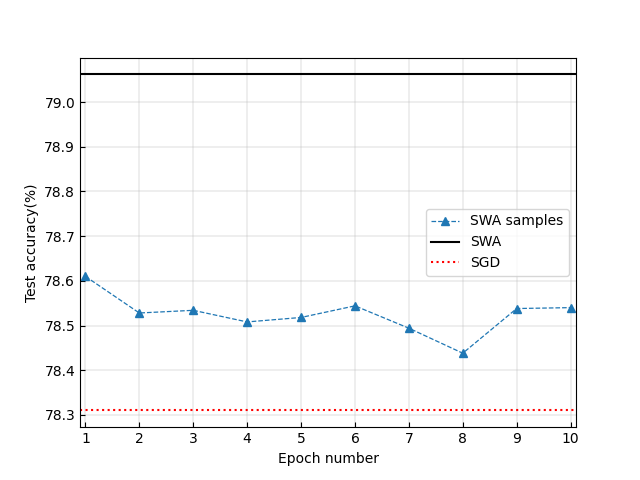}\includegraphics[width=0.5\linewidth]{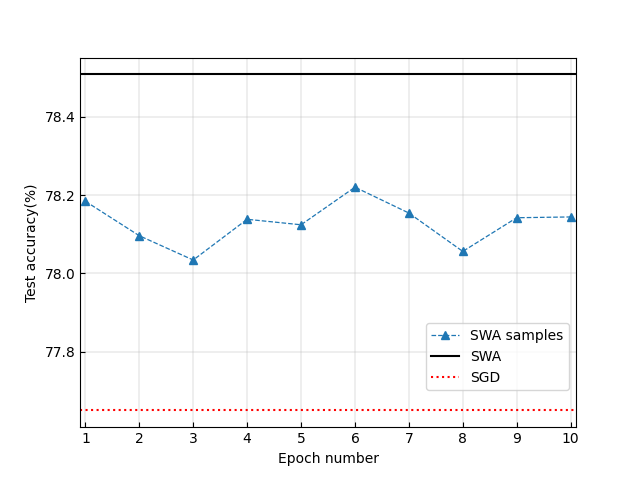}
\caption{Ablation study using the Imagenet dataset. The legends are defined in the same way as in Figure \ref{fig:ablation}. The top left, top right, bottom left and bottom right panels show results corresponding to VGG16, ResNet-50, ResNet-152, and DenseNet-161, respectively. Note that SWA begins based on such well pretrained models contained in Pytorch. So the horizontal axis label starts with epoch 1.}\label{fig:ablation3}
\end{figure}

We conduct the same ablation study as in Section \ref{sec:role} on the Imagenet dataset. We run SWA based on backbone DNN models VGG16, ResNet-50, ResNet-152, and DenseNet-161, which are contained in PyTorch. The results are presented in Figure \ref{fig:ablation3}.
\subsection{Experiments with a toy CNN model}\label{sec:exp_clean}
\begin{figure}[!htbp]
\centering
\includegraphics[width=0.5\linewidth]{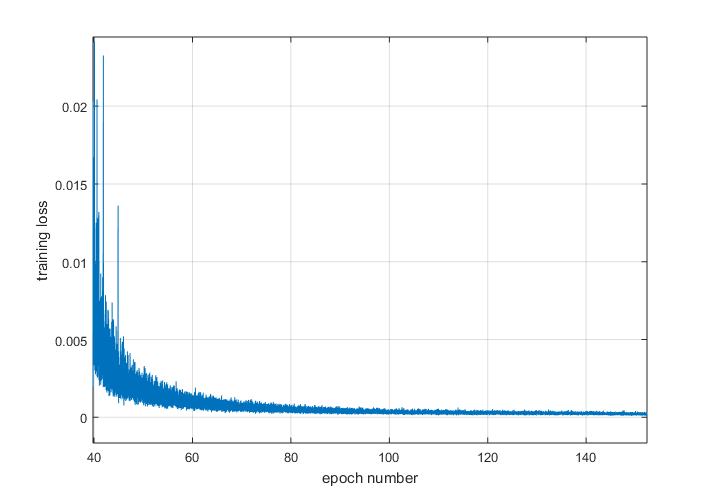}\includegraphics[width=0.5\linewidth]{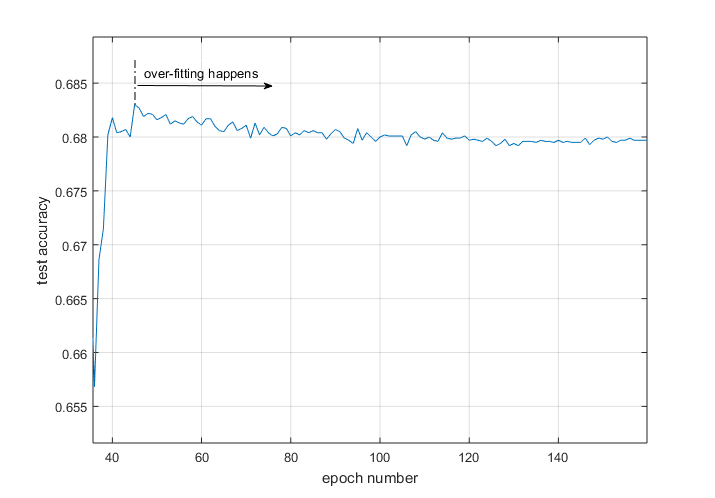}
\caption{The over-fitting result obtained when training a toy CNN model on CIFAR-10. This CNN has 9 layers: the input layer, the convolution layer, a max-pooling layer, another convolution layer, another max-pooling layer, the flatten layer, 2 fully connected layers, and a softmax layer.}\label{fig:cifar10_overfit}
\end{figure}

In this experiment, we remove the momentum module from SGD. We train a toy CNN model on CIFAR-10 and get an over-fitting result as shown in Figure \ref{fig:cifar10_overfit}. We collect the weight value at the end of each epoch and calculate its corresponding TA value. The maximum TA value of 0.683 appears at the 45th epoch. The TA corresponding to the last iterate of SGD is 0.680. We replace the last $L=5$ iterations of SGD with the SWA procedure, then get a TA value $w_{\mbox{swa}}=0.679$. We change the value of $L$ to be 20, getting $w_{\mbox{swa}}=0.680$. It is indicated that SWA does not lead to wider optima in this case.

\begin{figure}[!htbp]
\centering
\includegraphics[width=0.5\linewidth]{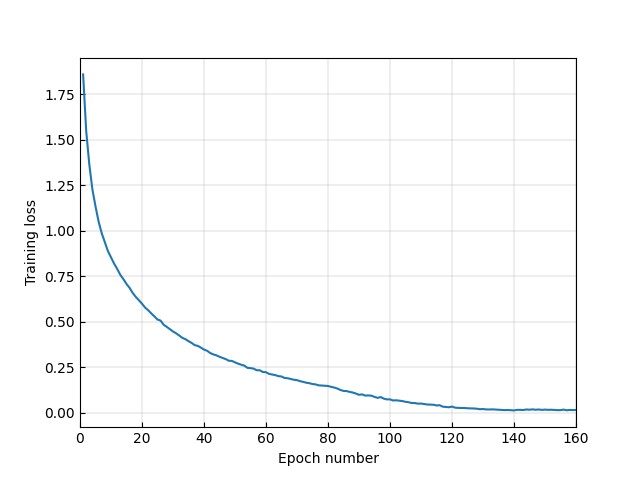}\includegraphics[width=0.5\linewidth]{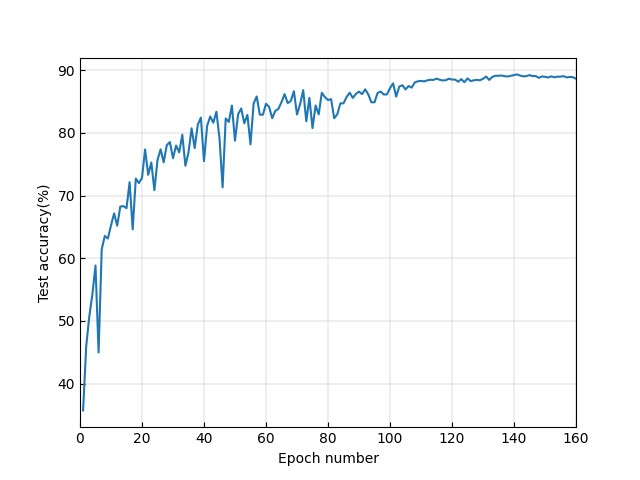}
\caption{Training PreResNet-164 on CIFAR-10.}\label{fig:cifar10_preresnet}
\end{figure}

A similar phenomenon happens when we replace the toy CNN model with PreResNet-164. We use the code open-sourced by \cite{izmailov2018averaging}, while closing off the momentum and the $L$2-based weight regularization to remove their effects on SWA. As is shown in Figure \ref{fig:cifar10_preresnet}, SGD converges after about the 120th epoch with TA achieving 89.24\%. We run SWA after the 140th epoch and get a TA of 89.17\%, which is smaller than the TA given by the converged SGD.
\subsection{Experiments with graph data}\label{sec:exp_graph}
We show experimental settings associated with the graph data experiments presented in Section \ref{sec:does} in Tables \ref{tab:parameter_GNN_exp}-\ref{tab:parameter_GNN_exp_PROTEINS}.
\begin{table}[!htbp]
\caption{The parameter setting for the GNN experiments. The baseline optimizer is Adam with weight decaying factor 0.0005. $L$ denotes the total number of epochs, $\alpha$ the learning rate of the Adam optimizer, $\alpha_{\mbox{SWA}}$ the constant learning rate used by SWA, and $t_{\mbox{SWA}}$ the starting point to launch SWA.}
\centering
\begin{tabular}{cccc}
\hline
parameter & GCN & GraphSAGE & GAT \\
\hline
$L$ & 200 & 20 & 300 \\
$\alpha$ & 0.01 & 0.003 & 0.005 \\
$t_{\mbox{SWA}}$ & 180 & 15 & 270 \\
$\alpha_{\mbox{SWA}}$ & 0.02 & 0.01 & 0.01 \\
\hline
\end{tabular}
\label{tab:parameter_GNN_exp}
\end{table}
\begin{table}[!htbp]
\caption{The parameter setting for the graph classification task on dataset NCI1. The baseline optimizer is Adam with weight decaying factor 0.0005. $L$ denotes the total number of epochs, $\alpha$ the learning rate of the Adam optimizer, $\alpha_{\mbox{SWA}}$ the constant learning rate used by SWA, and $t_{\mbox{SWA}}$ the starting point to launch SWA.}
\centering
\begin{tabular}{ccc}
\hline
parameter & MinCutPool & SAGPool \\
\hline
$L$ & 1000 & 300 \\
$\alpha$ & 0.001 & 0.003  \\
$t_{\mbox{SWA}}$ & 900 & 270  \\
$\alpha_{\mbox{SWA}}$ & 0.01 & 0.01 \\
\hline
\end{tabular}
\label{tab:parameter_GNN_exp_NCI1}
\end{table}
\begin{table}[!htbp]
\caption{The parameter setting for the graph classification task on dataset D\&D. The baseline optimizer is Adam with weight decaying factor 0.0005. $L$ denotes the total number of epochs, $\alpha$ the learning rate of the Adam optimizer, $\alpha_{\mbox{SWA}}$ the constant learning rate used by SWA, and $t_{\mbox{SWA}}$ the starting point to launch SWA.}
\centering
\begin{tabular}{ccc}
\hline
parameter & MinCutPool & SAGPool \\
\hline
$L$ & 50 & 150 \\
$\alpha$ & 0.001 & 0.003  \\
$t_{\mbox{SWA}}$ & 35 & 120  \\
$\alpha_{\mbox{SWA}}$ & 0.01 & 0.005 \\
\hline
\end{tabular}
\label{tab:parameter_GNN_exp_DD}
\end{table}
\begin{table}[H]
\caption{The parameter setting for the graph classification task on dataset PROTEINS. The baseline optimizer is Adam with weight decaying factor 0.0005. $L$ denotes the total number of epochs, $\alpha$ the learning rate of the Adam optimizer, $\alpha_{\mbox{SWA}}$ the constant learning rate used by SWA, and $t_{\mbox{SWA}}$ the starting point to launch SWA.}
\centering
\begin{tabular}{ccc}
\hline
parameter & MinCutPool & SAGPool \\
\hline
$L$ & 500 & 300 \\
$\alpha$ & 0.0001 & 0.003  \\
$t_{\mbox{SWA}}$ & 450 & 270  \\
$\alpha_{\mbox{SWA}}$ & 0.001 & 0.01 \\
\hline
\end{tabular}
\label{tab:parameter_GNN_exp_PROTEINS}
\end{table}
\end{document}